# POSSIBILITY AS SIMILARITY:
# THE SEMANTICS OF FUZZY LOGIC

Enrique H. Ruspini
Artificial Intelligence Center
SRI International
Menlo Park, California, U.S.A.*

*This work was supported in part by the US Air Force Office of Scientific Research under Contract No. F49620-89-K-0001 and in part by the USArmy Research Office under Contract No. DAAL03-89-K-0156. The views, opinions and/or conclusions contained in this note are those of the author and should not be interpreted as representative of the official positions, decisions, or policies, either express or implied, of the Air Force Office of Scientific Research, the Department of the Army, or the United States Government.

## Abstract

This paper addresses fundamental issues on the nature of the concepts and structures of fuzzy logic, focusing, in particular, on the conceptual and functional differences that exist between probabilistic and possibilistic approaches.

A semantic model provides the basic framework to define possibilistic structures and concepts by means of a function that quantifies proximity, closeness, or resemblance between pairs of possible worlds. The resulting model is a natural extension, based on multiple conceivability relations, of the modal logic concepts of necessity and possibility. By contrast, chance-oriented probabilistic concepts and structures rely on measures of set extension that quantify the proportion of possible worlds where a proposition is true.

Resemblance between possible worlds is quantified by a generalized similarity relation, i.e., a function that assigns a number between 0 and 1 to every pair of possible worlds. Using this similarity relation, which is a form of numerical complement of a classic metric or distance, it is possible to define and interpret the major constructs and methods of fuzzy logic: conditional and unconditional possibility and necessity distributions and the generalized modus ponens of Zadeh.

## 1 Introduction

In this paper, we present a semantic model of the major concepts, structures, and methods of fuzzy or possibilistic [13,14] logic. This model is based on a framework that combines the notion of a possible world [1], i.e., a potential state or situation of a real-world system, with measures of proximity or resemblance between pairs of possible worlds. The resulting structures are substantially different in character from those of probabilistic reasoning, based on measures of set extension, that quantify the proportion of possible worlds where a given proposition is true.

The results reported in this paper are the latest in a continuing investigative effort aimed at clarifying basic conceptual similarities and differences between a number of approaches to the treatment of imprecision and uncertainty. Using also possible-world semantic models, prior research has established that the Dempster-Shafer calculus of evidence may be interpreted by structures that result from the combination of conventional probabilistic calculus with epistemic logics [7]. By contrast, the formal structures discussed herein clearly show that fuzzy logic may be understood in a straightforward fashion by using conventional metric notions in a space of possible worlds without resorting in any form to probabilistic concepts. Furthermore, the actual functions that are used to combine possibilistic knowledge are substantially different from those used in the probability calculus.

Our exposition, which will be limited to the major structures of fuzzy logic, defines possibilistic concepts by using a more primive notion that has been found to be an essential component of important human cognitive processes [11]. The notion of similarity, in spite of its importance in reasoning processes, has not received substantial attention in treatments based on the use of logical concepts.

Perhaps as a consequence of its reliance on methods for the manipulation of symbolic strings and on a single partial order relation (implication) between formulas as the basis for its derivation procedures, there has been little attention given in formal logic to the consideration of other formal structures that capture important features of human knowledge such as the resemblance that exists between situations or circumstances. Although, for example, stating that the temperature is 28°C as opposed to saying that it is 29°C may be rather inconsequential in terms of the implication of either statement to a decision-maker (e.g., trying to decide what to wear), there is nothing in the basic framework of logic that makes the second statement any more different than to say that the temperature is below freezing (as neither of the three statements is logically consistent with the other two).

The determination and use of similarity information is, however, not only central to all forms of analogical reasoning but it is an essential element in the derivation of physical law. Formal studies in measurement theory [6] clearly show the role that measures based on similar behavior play in the derivation of rational measurement schemes, while also explaining the ubiquitous presence of numeric scales throughout science.

The results presented in this paper show that, when such notions of proximity are formalized in the context of a possible-worlds model, the major functional structures of fuzzy logic, possibility and necessity distributions, and its major inferential procedure, the generalized modus ponens of Zadeh, may be readily explained as a natural extension of classical logical concepts. In particular, possibility and necessity distributions simply correspond to best and worst scenarios in a space of possible real-world states, while the generalized modus ponens [14] is a sound inferential procedure that may be regarded as a form of logical extrapolation between neighboring situations.

The scope of this paper prevents a detailed discussion of all pertinent results and derivations. A complete account of all relevant matters regarding this similarity-based model of fuzzy logic is presented in a related technical note [8], which this paper summarizes.

## 2 The Approximate Reasoning Problem

Our model of the approximate reasoning problem is based on the notion of "possible world." Informally, possible worlds are the conceivable states of affairs of a real-world system that are consistent with the laws of logic.

Restricting ourselves, for the sake of simplicity to propositional formulations, a possible world is a function [1] that assigns a unique conventional truth value (i.e., true or false) to every proposition that describes some relevant aspect of the state of a real-world system and that, in addition, satisfies the axioms of propositional logic.

In the absence of any knowledge about the behavior of a system of interest or of any observation about its state, it is impossible to determine which, among all conceivable situations, corresponds to the actual state of the real world. Availability of factual evidence or determination of the laws of behavior of the system permits us, however, to eliminate from consideration some possible worlds in this *universe of discourse*. The remaining possible worlds correspond to satisfiable propositions that, in addition, are logically consistent with the evidence. This subset of conceivable situations or scenarios will be called the *evidential set*, denoted $\mathscr{E}$.

If the typical reasoning problem is thought of as the determination of the truth value of a proposition $h$ (the hypothesis), then an approximate reasoning problem may be described as one where available evidence does not permit such evaluation without ambiguity. In other words, as illustrated in Figure 1, there are some members of the evidential set where the hypothesis is true and some where it is false.

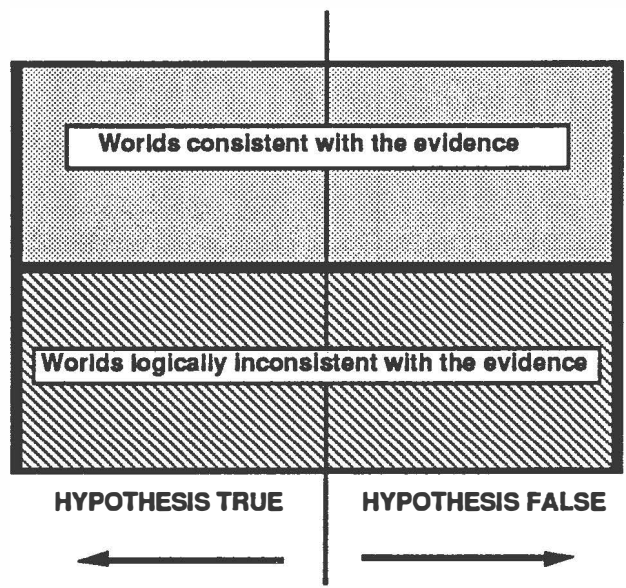

Figure 1: The approximate reasoning problem

Our approach to the formalization of the major concepts and structures of fuzzy logic is based on a generalization of a central concept of semantic models of modal logics. Modal logics [3] may be generally described as extensions of conventional two-valued logic that permit us to qualify, in various ways, the meaning of propositional truth.

In our model, we utilize modal concepts to explain basic possibilistic structures using the more primitive notion of similarity. This notion is introduced, however, by means of conventional set-theoretic and logical concepts. In this regard, our approach to the study of the interplay of modal and possibilistic logics is different from approaches such as that used by Lakoff [5], who sought to generalize modal logics using fuzzy-set concepts; or that of Dubois and Prade [2], who investigated modal structures with a view to the development of formal proof mechanisms in possibilistic logic.

A major concept of semantic models of modal logic systems is a binary relation $R$, called the *accessibility* or *conceivability* relation, which is assumed to have a number of properties intended to capture the semantics of various qualifications of propositional truth.

Our aim is to characterize the extent to which statements that are true in one situation or scenario may be said, perhaps with some suitable modification, to be true in another state of affairs that resembles the first. We are particularly interested in describing more general (i.e., less specific) propositions that are true in one possible world as a function of the propositions that are true in another. In order to model a continuous range of proximity between possible worlds, we will generalize the notion of accessibility relation to a full family of binary relations $R_\alpha$, indexed by a numerical parameter $\alpha$ taking values between 0 and 1, along the same lines, albeit with a different purpose, as Lewis in his treatment of counterfactuals [4].

## 3 Similarity Functions

We will introduce a family of accessibility relations

$$\{R_\alpha : \alpha \text{ in } [0,1]\},$$

by means of a binary function $S$, called the *similarity relation*, that maps pairs of possible worlds into numbers between 0 and 1. The multiple relations of accessibility $R_\alpha$ are defined in terms of this similarity function by

$$wR_\alpha w', \text{ whenever } S(w, w') \geq \alpha.$$

The function $S$ is intended to capture a notion of proximity, closeness, or resemblance between possible worlds, with a value of 1 corresponding to the identity of possible worlds and a value of 0 indicating that knowledge of propositions that are true in one possible world does not provide any indication about propositions that are true in the other. To assure that the function $S$ has the semantics of a relation that quantifies resemblance between possible states of affairs it is necessary to require that it satisfies a number of properties.

Besides the above mentioned property that the similarity between a possible world and itself has the highest possible value, equivalent to stating that each accessibility relation $R_\alpha$ is reflexive, we will also require that the similarity between different possible worlds be strictly less than one. This requirement is intended to assure that the similarity relation may distinguish between different states of the possible world.

The similarity relation will also be assumed to be symmetric, and to satisfy a relaxed form of transitivity. Clearly, if the pairs of possible worlds $(w, w')$ and $(w', w'')$ correspond to highly similar situations, it would be surprising if $w$ and $w''$ were highly dissimilar. It is natural to assume, therefore, that

$$S(w, w'') \geq S(w, w') \circledast S(w', w''),$$

where $\circledast$ is a binary operator used to represent the lower bound for $S(w, w'')$ as a function of its arguments. This requirement is equivalent to the relaxed transitivity condition

$$R_{\alpha \circledast \beta} \subseteq R_\alpha \circ R_\beta,$$

which replaces the usual, more stringent, definition of transitivity.

Imposition of reasonable requirements upon the function $\circledast$ shows that it has the properties of a triangular norm [9]. These functions, which play a significant role in multivalued logics [10], may be justified, therefore, purely on the basis of metric considerations. Important examples of triangular norms are

$$a \circledast b = \min(a, b), \ a \circledast b = \max(a+b-1, 0),$$

and

$$a \circledast b = ab,$$

called the *Zadeh*, *Lukasiewicz*, and *product* triangular norms, respectively.

The generalized transitivity property expressed by triangular norms clarifies their relationship to the conventional mathematical concept of metric. If $S$ is a similarity function, then the function $\delta = 1 - S$ has the properties of a distance function. When $\circledast$ corresponds to the Lukasiewicz norm, then the transitivity property of $S$ corresponds to the well-known triangular property of distance functions. If $\circledast$ correponds to the Zadeh triangular norm, then $\delta$ may be shown to satisfy the more stringent ultrametric inequality.

## 4 Degrees of Implication and Consistence

The classical derivation rule of modus ponens is easily interpreted in terms of subsets of possible worlds, since each proposition corresponds to one such set, while implication corresponds to the conventional relation of subset inclusion. In short, if three propositions $p$, $q$ and $r$ are equated with the corresponding subsets of possible worlds, and if it is known that $p$ is a subset of $q$ (i.e., $p \Rightarrow q$) and that $q$ is a subset of $r$ (i.e., $q \Rightarrow r$), then the modus ponens simply states that $p$ is a subset of $r$ (i.e., $p \Rightarrow r$).

While only two degrees of inclusion are recognized in two-valued logic, in that a set is either a subset of another set or it is not, the introduction of a similarity structure permits the definition of a graded notion of inclusion.

To illustrate the relationship between this new concept and generalized modal concepts, we will extend first the classical definition of the modal operator $\Pi$ in the following way: a proposition $p$ will be said to be *possible to the degree* $\alpha$ in the possible world $w$ (denoted $w \vdash \Pi_\alpha p$) if there exists a world $w'$ such that $p$ is true in $w'$ and such that $S(w, w') \geq \alpha$. Furthermore, we will say that $q$ *necessarily implies*, or is a *necessary model of $p$ to the degree* $\alpha$, if every $q$-world $w$ is such that $p$ is possible to the degree $\alpha$ in $w$, i.e.,

$$q \Rightarrow \Pi_\alpha p.$$

The definition of graded necessary implication generalizes the corresponding modal notion, by permitting us to express that $q$ is a subset of a neighborhood, of specified size, of $p$. Since every possible world is similar, at least to the degree zero, to every other pos-sible world, it is obvious that every subset of possible worlds implies every other subset to some degree $\alpha$ between 0 and 1. In simpler words, if $p$ is stretched[1] to the degree $\alpha$, then this stretched set will include $q$. The upper bound of the values $\alpha$ such that $q$ necessarily implies $p$ to the degree $\alpha$, expressed by

$$\mathbf{I}(p\,|\,q) = \inf_{w' \vdash q} \sup_{w \vdash p} S(w, w'),$$

provides a most useful measure of the minimal amount of stretching required to cover $q$ with a neighborhood of $p$, as illustrated in Figure 2. The function $\mathbf{I}$ will be called the *degree of implication*.

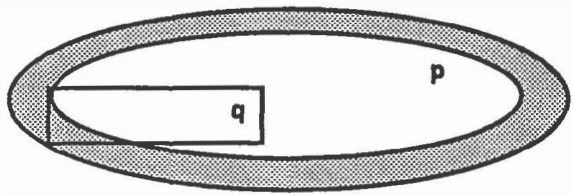

Figure 2: Degree of implication

The degree of implication function provides the basis for the derivation of the generalized modus ponens of Zadeh [13]. This rule generalizes its classical counterparts by stating, in quantitative terms, the inclusion relations that hold between neighborhoods of three propositions, which is expressed by the transitive property of the degree of implication function

$$\mathbf{I}(p\,|\,q) \geq \mathbf{I}(p\,|\,r) \circledast \mathbf{I}(r\,|\,q).$$

This transitive property may also be expressed using modal operators in the form

$$q \Rightarrow \Pi_\alpha r \quad \text{and} \quad r \Rightarrow \Pi_\beta p$$

---

[1] Recall that $S$ is a similarity measure and that lower values of $\alpha$ correspond to higher values of stretching.

imply that

$$q \Rightarrow \Pi_{\alpha \circledast \beta} p.$$

This simplest form of the generalized modus ponens, illustrated in Figure 3, tells us how much $p$ should be stretched to encompass $q$ on the basis of knowledge of the sizes of the neighborhoods of $p$ that includes $r$ and of $r$ that includes $q$.

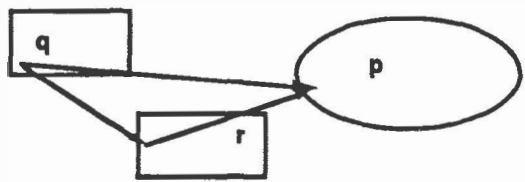

Figure 3: The generalized modus ponens

In addition to a function defining a degree of implication, we will also need a dual notion, called the *degree of consistency*, to quantify the amount by which a set must be stretched in order to intersect another set. The *degree of consistency* of $p$ and $q$ is defined by the expression

$$\mathbf{C}(p\,|\,q) = \sup_{w' \vdash q} \sup_{w \vdash p} S(w, w').$$

It is obvious from the definition that

$$\mathbf{I}(p\,|\,q) \leq \mathbf{C}(p\,|\,q).$$

## 5 Possibilistic Distributions

The customary expression of the generalized modus ponens in fuzzy logic, which is more general than the simple transitive law of the degree of implication given above, is expressed by means of possibility and necessity distributions. As is the case with their probabilistic counterparts, these functions may be defined in both unconditioned and conditional forms, corresponding, respectively, to descriptions of the proximity of the evidential set to other subsets of possible worlds with or without additional constraints.

The unconditioned necessity of a set is defined as any lower bound of the degree of implication of $p$ by $\mathscr{E}$; thus $\mathbf{Nec}(\cdot)$ is called an *unconditioned necessity distribution* for $\mathscr{E}$ if

$$\mathbf{Nec}(p) \leq \mathbf{I}(p\,|\,\mathscr{E}).$$

The corresponding concept of unconditioned possibility distribution is dually defined as any upper bound for the degree of consistence of $p$ and $\mathscr{E}$, i.e., $\mathbf{Poss}(\cdot)$ is called an *unconditioned possibility distribution* for $\mathscr{E}$ if

$$\mathbf{Poss}(p) \geq \mathbf{C}(p\,|\,\mathscr{E}).$$

This interpretation of necessity and possibility distributions readily leads to the derivation of a number of properties for these functions such as the basic laws, usually assumed as axioms, linking the possibility and necessity of disjunctions with those of the disjuncts.

To define conditional possibility and necessity distributions it is necessary to introduce a form of inverse of the triangular norm. This function permits us to state the transitive property of similarity relations by using an equivalent formulation of triangular inequalities, along the lines used in plane geometry to express the relation that exists between the three sides of a triangle, i.e., either (the length of) a side is smaller than the sum of the other two, or a side is larger than the difference of the other two.

If $\circledast$ is a triangular norm, its *pseudoinverse* $\oslash$ is the function, defined over pairs of numbers in the unit interval of the real line, by the expression

$$a \oslash b = \sup\{\,c :\ b \circledast c \leq a\,\}.$$

The following definitions make use of the inverse of triangular norms to characterize conditional possibility and necessity distributions as measures of the proximity relations that exist between evidential worlds and those worlds satisfying a consequent proposition $q$ as a proportion of the similarity that exists between those evidential worlds and worlds that satisfy the antecedent proposition $p$.

**Definition:** A function $\mathbf{Nec}(\cdot|\cdot)$ is called a *conditional necessity distribution* for $\mathscr{E}$ if

$$\mathbf{Nec}(q|p) \leq \inf_{w \vdash \mathscr{E}} [\,\mathbf{I}(q\,|\,w) \oslash \mathbf{I}(p\,|\,w)\,].,$$

**Definition:** A function $\mathbf{Poss}(\cdot|\cdot)$ is called a *conditional possibility distribution* for $\mathscr{E}$ if

$$\mathbf{Poss}(q|p) \geq \sup_{w \vdash \mathscr{E}} [\,\mathbf{I}(q\,|\,w) \oslash \mathbf{I}(p\,|\,w)\,].$$

## 6 Generalized Modus Ponens

The *compositional rule of inference* or generalized modus ponens of of Zadeh [13] generalizes the corresponding classical rule of inference, permitting to infer statements even when a perfect match does not exist between the antecedent of a conditional rule and known facts, i.e., from

$$\frac{\begin{array}{c} p \\ p \;\rightarrow\; q \end{array}}{q}$$

to its approximate version

$$\frac{\begin{array}{c} p' \\ p \;\rightarrow\; q \end{array}}{q'}$$

where $p'$ and $q'$ are similar to $p$ and $q$, respectively.

The generalized modus ponens may be viewed, therefore, as an extrapolation procedure that exploits the similarity between the known facts, or the evidence formalized as the evidential set, and a set of possible worlds or the antecedent proposition $p$, together with expressions relating the proximity of $p$-worlds and $q$-worlds, to bound the similarity between evidential worlds and those that satisfy the consequent proposition $q$.

The actual formal statement of the generalized modus ponens makes use of the notion of partition of the universe of discourse. As a simple extension of the notion of disjoint partition of a set, a *partition* $\mathscr{P}$ is simply defined as a collection of mutually inconsistent propositions such that their disjunction is always true, or, equivalently, as a collection of disjoint subsets of possible worlds such that their union is the universe of discourse. In most applications of fuzzy set theory, partitions follow naturally from the notion of variable since every possible world corresponds to a unique value of a system variable.

Using this concept, it is possible to state the following results that allow derivation of necessity and possibility distributions for a consequent proposition $q$, on the basis of approximate knowledge expressed by unconditioned distributions for antecedent propositions $p$ in the partition $\mathscr{P}$, and conditional distributions relating these antecedent propositions with $q$.

**Theorem** (*Generalized Modus Ponens for Necessity Functions*): Let $\mathscr{P}$ be a partition and let $q$ be a proposition. If $\mathbf{Nec}(p)$ and $\mathbf{Nec}(q|p)$ are real values, defined for every proposition $p$ in the partition $\mathscr{P}$, such that

$$\mathbf{Nec}(p) \leq \mathbf{I}(p|\mathscr{E}),$$
$$\mathbf{Nec}(q|p) \leq \inf_{w \vdash \mathscr{E}} [\mathbf{I}(q|w) \oslash \mathbf{I}(p|w)],$$

then the following inequality is valid:

$$\sup_{\mathscr{P}} [\mathbf{Nec}(q|p) \circledast \mathbf{Nec}(p)] \leq \mathbf{I}(q|\mathscr{E}).$$

**Theorem** (*Generalized Modus Ponens for Possibility Functions*): Let $\mathscr{P}$ be a partition and let $q$ be a proposition. If $\mathbf{Poss}(p)$ and $\mathbf{Poss}(q|p)$ are real values, defined for every proposition $p$ in $\mathscr{P}$, such that

$$\mathbf{Poss}(p) \geq \mathbf{C}(p|\mathscr{E}),$$
$$\mathbf{Poss}(q|p) \geq \sup_{w \vdash \mathscr{E}} [\mathbf{I}(q|w) \oslash \mathbf{I}(p|w)],$$

then the following inequality is valid:

$$\sup_{\mathscr{P}} [\mathbf{Poss}(q|p) \circledast \mathbf{Poss}(p)] \geq \mathbf{C}(q|\mathscr{E}).$$

## 7 Conclusion

This paper summarizes some aspects of a semantic model of fuzzy logic that is rooted on notions of similarity or resemblance between conceivable states of the real world. These metric structures are conceptually and functionally different from those that are used in probabilistic reasoning, which rely on measures of set extension.

The essential difference between the aims and structures of probabilistic and possibilistic formulations that is clarified by this model is readily understood when it is noted that being in a situation $s$ does not make a similar situation $s'$ more likely, or vice versa. Reality is described in one case by summary accounts of past experience and, in the other, by comparison with other states introduced for referential purposes, regardless of their degree of historic ubiquity.

The insight provided by this model makes clear that possibilistic notions describe situations known through imprecise, uncertain, and vague information in a way that neither replaces or is replaced but that, rather, complements the views produced by other approaches.